\documentclass[11pt]{article}
\usepackage{amssymb}
\usepackage{amsmath}
\usepackage{amsfonts}
 \usepackage{pifont}
\usepackage{amsmath,amsfonts,amssymb,theorem,graphicx,listings,txfonts}
\usepackage{graphics}
 \usepackage{caption2}
\usepackage{flafter}
 \usepackage{indentfirst}
\usepackage{epsfig}
  \usepackage{mathrsfs}
\usepackage{subfigure}
 \usepackage{cite}
\newtheorem{theorem}{Theorem}[section]

\newtheorem{definition}[theorem]{Definition}
\newtheorem{algorithm}[theorem]{Algorithm}

\theoremstyle{example}
\newtheorem{example}[theorem]{Example}
\theoremstyle{programme}

\theoremstyle{property}

\theoremstyle{problem}

\title{Knowledge reduction of dynamic covering decision information systems with immigration of more objects}

\author
{Guangming Lang
\thanks{Corresponding author.\quad Tel./fax: +86 731 85258787,
langguangming1984@126.com
\newline\mbox{}\hspace{0.55cm}
. }\hspace{1cm}
\\
\small { School of Mathematics and Computer Science, Changsha University of Science and Technology}\\
\small {Changsha, Hunan 410114, P.R. China}\\
}
\date{}

\begin{document}
\maketitle \baselineskip=17pt
\begin{center}
\begin{quote}
{{\bf Abstract.}
In practical situations, it is of interest to investigate computing
approximations of sets as an important step of knowledge reduction
of dynamic covering decision information systems. In this paper, we present
incremental approaches to computing the type-1 and type-2
characteristic matrices of dynamic coverings whose cardinalities increase with immigration of more objects. We also present the incremental algorithms of computing the second and sixth lower and upper approximations of sets in dynamic covering approximation spaces. 

\\}
\end{quote}
\end{center}
\renewcommand{\thesection}{\arabic{section}}

\section{Introduction}

Covering-based rough set theory\cite{Zakowski1}, as a powerful
mathematical tool for studying covering approximation spaces, has
attracted a lot of attention of researchers in various fields of
sciences. Especially, various kinds of approximation operators have
been proposed for covering approximation spaces. Recently, Wang et
al.\cite{Wang1} transformed the computation of approximations of a
set into products of the characteristic matrices and the
characteristic function of the set. However, it paid little
attention to approaches to calculating the characteristic matrices.
In practice, the covering approximation space varies with time due
to the characteristics of data collection, and the non-incremental
approach to constructing the characteristic matrices is often very
costly or even intractable in dynamic covering approximation spaces.
It is necessary to present effective approaches to computing
characteristic matrices of dynamic coverings.

To the best of our knowledge,
researchers\cite{Chen1,Chen3,Li12,Luo1,Luo2, Zhang1,Zhang4} have
focused on computing approximations of sets. For instance, Chen et
al.\cite{Chen1,Chen3} constructed approximations of sets when
coarsening or refining attribute values. Li et al.\cite{Li12}
computed approximations in dominance-based rough sets approach under
the variation of attribute set. Luo et al.\cite{Luo1,Luo2} studied
dynamic maintenance of approximations in set-valued ordered decision
systems under the attribute generalization and the variation of
object set. Zhang et al.\cite{Zhang1,Zhang4} updated rough set
approximations based on relation matrices and investigated
neighborhood rough sets for dynamic data mining. These works
demonstrate that incremental approaches are effective and efficient
for computing approximations of sets. It motivates us to apply an
incremental updating scheme to conduct approximations of sets by
using characteristic matrices in dynamic covering approximation
spaces, which will provide an effective approach to computing
approximations of sets from the view of matrices.

The purpose of this paper is to compute approximations of sets by
using incremental approaches in dynamic covering approximation
spaces. First, we present incremental approaches to computing the
type-1 and type-2 characteristic matrices in dynamic covering
approximation spaces. We mainly focus on the situation: the
variation of elements in coverings when adding and deleting objects.
Furthermore, we provide incremental algorithms for constructing the
second and sixth lower and upper approximations of sets based on the
type-1 and type-2 characteristic matrices, respectively. We compare
computation complexities of the incremental algorithms with those of
non-incremental algorithms. Several examples are employed to
illustrate that calculating approximations of sets is simplified
greatly by utilizing the proposed approach.

The rest of this paper is organized as follows: Section 2 briefly
reviews the basic concepts of covering-based rough set theory. In
Section 3, we introduce incremental approaches to computing the
type-1 and type-2 characteristic matrices of dynamic coverings whose cardinalities increase with immigration of more objects. In Section 4, we present incremental algorithms of calculating the
second and sixth lower and upper approximations of sets by using the
type-1 and type-2 characteristic matrices, respectively. We conclude the paper in Section 5.

\section{Preliminaries}

In this section, we  briefly review some concepts of covering-based rough
sets.

\begin{definition}\cite{Zakowski1}
Let $U$ be a finite universe of discourse, and $\mathscr{C}$ a
family of subsets of $U$. Then $\mathscr{C}$ is called a covering of
$U$ if none of elements of $\mathscr{C}$ is empty and
$\bigcup\{C|C\in \mathscr{C}\}=U$.
\end{definition}

If $\mathscr{C}$ is a covering of $U$, then $(U,\mathscr{C})$ is referred to as a covering approximation
space.

\begin{definition}\cite{Wang1}
Let $U=\{x_{1},x_{2},...,x_{n}\}$ be a finite universe, and
$\mathscr{C}=\{C_{1},C_{2},...,C_{m}\}$ a covering of $U$. For any
$X\subseteq U$, the second, fifth and sixth upper and lower
approximations of $X$ with respect to $\mathscr{C}$
are defined as

$(1)$ $SH_{\mathscr{C}}(X)=\bigcup\{C\in\mathscr{C}|C\cap X\neq
\emptyset\}$, $SL_{\mathscr{C}}(X)=[SH_{\mathscr{C}}(X^{c})]^{c}$;

$(2)$ $IH_{\mathscr{C}}(X)=\bigcup\{N(x)|N(x)\cap X\neq \emptyset,x\in U\}$,
$IL_{\mathscr{C}}(X)=\bigcup\{N(x)|N(x)\subseteq X,x\in U\}$;

$(3)$ $XH_{\mathscr{C}}(X)=\{x\in U|N(x)\cap X\neq \emptyset\}$,
$XL_{\mathscr{C}}(X)=\{x\in U|N(x)\subseteq X\}$,
 where $N(x)=\bigcap\{C_{i}|x\in C_{i}\in \mathscr{C}\}$.
\end{definition}

For simplicity, we omit $\mathscr{C}$ in the following
description of approximation operators.

\begin{definition}\cite{Wang1}
Let $U=\{x_{1},x_{2},...,x_{n}\}$ be a finite universe,
$\mathscr{C}=\{C_{1}, C_{2}, ..., C_{m}\}$ a family of subsets of $U$,
and $M_{\mathscr{C}}=(a_{ij})_{n\times m}$, where $a_{ij}=\left\{
\begin{array}{ccc}
1,&{\rm}& x_{i}\in C_{j};\\
0,&{\rm}& x_{i}\notin C_{j}.
\end{array}
\right. $ Then $M_{\mathscr{C}}$ is called a matrice representation
of $\mathscr{C}$.
\end{definition}

Accordingly, we have the characteristic function $\mathcal
{X}_{X} =\left[\begin{array}{cccccc}
  a_{1}&a_{2}&.&.&. & a_{n} \\
  \end{array}
\right]^{T}$ for $X\subseteq U$, where $a_{i}=\left\{
\begin{array}{ccc}
1,&{\rm}& x_{i}\in X;\\
0,&{\rm}& x_{i}\notin X.
\end{array}
\right. $.

\begin{definition}\cite{Wang1}
Let $\mathscr{C}$ be a covering of the
universe $U$, $A=(a_{ij})_{n\times m}$ and $B=(b_{ij})_{m\times p}$ Boolean
matrices, $A\odot B=(c_{ij})_{n\times p}$, where
$c_{ij}=\bigwedge^{m}_{k=1}(b_{kj}-a_{ik}+1).$ Then

$(1)$ $\Gamma(\mathscr{C})=M_{\mathscr{C}}\cdot
M_{\mathscr{C}}^{T}=(d_{ij})_{n\times n}$ is called the type-1 characteristic matrice
of $\mathscr{C}$, where $d_{ij}=\bigvee^{m}_{k=1}(a_{ik}\cdot
a_{jk})$, and $M_{\mathscr{C}}\cdot M_{\mathscr{C}}^{T}$ is the
boolean product of $M_{\mathscr{C}}$ and its transpose $
M_{\mathscr{C}}^{T}$;

$(2)$ $\prod(\mathscr{C})=M_{\mathscr{C}}\odot M_{\mathscr{C}}^{T}$ is referred to as the type-2 characteristic matrice of
$\mathscr{C}$.
\end{definition}

Wang et al. axiomatized two important types of covering
approximation operators equivalently by using the type-1 and  type-2 characteristic matrice of
$\mathscr{C}$.

\begin{definition}\cite{Wang1}
Let $U=\{x_{1},x_{2},...,x_{n}\}$ be a finite universe,
$\mathscr{C}=\{C_{1},C_{2},...,C_{m}\}$ a covering of $U$, and
$\mathcal {X}_{X}$ the characteristic function of $X$ in $U$. Then

$(1)$ $\mathcal {X}_{SH(X)}=\Gamma(\mathscr{C})\cdot \mathcal
{X}_{X}$, $\mathcal {X}_{SL(X)}=\Gamma(\mathscr{C})\odot \mathcal
{X}_{X}$; $(2)$ $\mathcal {X}_{XH(X)}=\prod(\mathscr{C})\cdot
\mathcal {X}_{X}$, $\mathcal {X}_{XL(X)}=\prod(\mathscr{C})\odot
\mathcal {X}_{X}$.
\end{definition}

\section{Update approximations of sets
with immigration of more objects}

In this section, we introduce incremental approaches to computing
the second and sixth lower and upper approximation of sets with immigration of more objects.

\begin{definition}
Let $(U,\mathscr{C})$ and $(U^{+},\mathscr{C}^{+})$ be covering
approximation spaces, where $U=\{x_{1},x_{2},...,x_{n}\}$,
$U^{+}=U\cup \{x_{n+1},x_{n+2},...,x_{n+t}\}(t\geq 2)$, $\mathscr{C}=\{C_{1},C_{2},...,C_{m}\}$,
$\mathscr{C}^{+}=\{C^{+}_{1},C^{+}_{2},...,C^{+}_{m},C^{+}_{m+1},C^{+}_{m+2},...,C^{+}_{m+l}\}(l\geq 2)$,
where $C^{+}_{i}=C_{i}\cup \Delta C_{i}$ or $C_{i}$ $(1\leq i\leq
m)$, $\Delta C_{i}\subseteq \{x_{n+1},x_{n+2},...,x_{n+t}\}$, and $\{x_{n+1},x_{n+2},...,x_{n+t}\}\subseteq \{C^{+}_{m+j}|1\leq j\leq l\}$. Then $(U^{+},\mathscr{C}^{+})$ is
called a dynamic covering approximation space.
\end{definition}

By Definition 3.1, we refer $\mathscr{C}^{+}$ to as a dynamic covering.
Although there are several types of coverings when adding
objects, we only discuss this type of dynamic coverings for simplicity in this work.

In what follows, we discuss how to construct
$\Gamma(\mathscr{C}^{+})$ based on $\Gamma(\mathscr{C})$. For
convenience, we denote $M_{\mathscr{C}}=(a_{ij})_{n\times m}$,
$M_{\mathscr{C}^{+}}=(a_{ij})_{(n+t)\times (m+l)}$,
$\Gamma(\mathscr{C})=(b_{ij})_{n\times n}$ and
$\Gamma(\mathscr{C}^{+})=(c_{ij})_{(n+t)\times (n+t)}$.

\begin{theorem}
Let $(U^{+},\mathscr{C}^{+})$ be a dynamic covering approximation space
of $(U,\mathscr{C})$, $\Gamma(\mathscr{C})$ and
$\Gamma(\mathscr{C}^{+})$ the type-1 characteristic matrices of
$\mathscr{C}$ and $\mathscr{C}^{+}$, respectively. Then
 \begin{eqnarray*}
 \Gamma(\mathscr{C}^{+})=\left[
  \begin{array}{cc}
    \Gamma(\mathscr{C}) & 0\\
    0 & 0\\
  \end{array}
\right]\bigvee \left[
  \begin{array}{cc}
    \triangle_{1}(\Gamma(\mathscr{C})) & (\triangle_{2}(\Gamma(\mathscr{C})))^{T}\\
    \triangle_{2}(\Gamma(\mathscr{C})) & \triangle_{3}(\Gamma(\mathscr{C}))\\
  \end{array}
\right],
\end{eqnarray*} where
  \begin{eqnarray*}
  \triangle_{1}(\Gamma(\mathscr{C}))&=&\left[\begin{array}{cccccc}
  a_{1(m+1)}&a_{2(m+1)}&.&.&. & a_{n(m+1)}\\
  a_{1(m+2)}&a_{2(m+2)}&.&.&. & a_{n(m+2)}\\
  .&.&.&.&. & .\\
  .&.&.&.&. & .\\
  .&.&.&.&. & .\\
  a_{1(m+l)}&a_{2(m+l)}&.&.&. & a_{n(m+l)}\\
  \end{array}
\right]^{T}\cdot \left[\begin{array}{cccccc}
  a_{1(m+1)}&a_{2(m+1)}&.&.&. & a_{n(m+1)}\\
  a_{1(m+2)}&a_{2(m+2)}&.&.&. & a_{n(m+2)}\\
  .&.&.&.&. & .\\
  .&.&.&.&. & .\\
  .&.&.&.&. & .\\
  a_{1(m+l)}&a_{2(m+l)}&.&.&. & a_{n(m+l)}\\
  \end{array}
\right];
 \\
\triangle_{2}(\Gamma(\mathscr{C}))&=&\left[\begin{array}{cccccc}
  a_{(n+1)1}&a_{(n+1)2}&.&.&. & a_{(n+1)(m+l)}\\
  a_{(n+2)1}&a_{(n+2)2}&.&.&. & a_{(n+2)(m+l)}\\
  .&.&.&.&. & .\\
  .&.&.&.&. & .\\
  .&.&.&.&. & .\\
  a_{(n+t)1}&a_{(n+t)2}&.&.&. & a_{(n+t)(m+l)}\\
  \end{array}
\right]\cdot \left[\begin{array}{cccccc}
  a_{11}&a_{12}&.&.&. & a_{1(m+l)} \\
  a_{21}&a_{22}&.&.&. & a_{2(m+l)} \\
  .&.&.&.&. & . \\
  .&.&.&.&. & . \\
  .&.&.&.&. & . \\
  a_{n1}&a_{n2}&.&.&. & a_{n(m+l)} \\
  \end{array}
\right]^{T};\\
 \triangle_{3}(\Gamma(\mathscr{C}))&=&\left[\begin{array}{cccccc}
  a_{(n+1)1}&a_{(n+1)2}&.&.&. & a_{(n+1)(m+l)}\\
  a_{(n+2)1}&a_{(n+2)2}&.&.&. & a_{(n+2)(m+l)}\\
  .&.&.&.&. & .\\
  .&.&.&.&. & .\\
  .&.&.&.&. &.\\
  a_{(n+t)1}&a_{(n+t)2}&.&.&. & a_{(n+t)(m+l)}\\
  \end{array}
\right]\cdot \left[\begin{array}{cccccc}
  a_{(n+1)1}&a_{(n+1)2}&.&.&. & a_{(n+1)(m+l)}\\
  a_{(n+2)1}&a_{(n+2)2}&.&.&. & a_{(n+2)(m+l)}\\
  .&.&.&.&. & .\\
  .&.&.&.&. & .\\
  .&.&.&.&. &.\\
  a_{(n+t)1}&a_{(n+t)2}&.&.&. & a_{(n+t)(m+l)}\\
  \end{array}
\right]^{T}.
\end{eqnarray*}
\end{theorem}

\noindent\textbf{Proof.} By Definition 3.1, we get
$\Gamma(\mathscr{C})$ and $\Gamma(\mathscr{C}^{+})$ as follows:
\begin{eqnarray*}
\Gamma(\mathscr{C})&=&M_{\mathscr{C}}\cdot
M_{\mathscr{C}}^{T}\\&=&\left[
  \begin{array}{cccccc}
    a_{11} & a_{12} & . & . & . & a_{1m} \\
    a_{21} & a_{22} & . & . & . & a_{2m} \\
    . & . & . & . & . & . \\
    . & . & . & . & . & . \\
    . & . & . & . & . & . \\
    a_{n1} & a_{n2} & . & . & . & a_{nm} \\
  \end{array}
\right] \cdot \left[
  \begin{array}{cccccc}
    a_{11} & a_{12} & . & . & . & a_{1m} \\
    a_{21} & a_{22} & . & . & . & a_{2m} \\
    . & . & . & . & . & . \\
    . & . & . & . & . & . \\
    . & . & . & . & . & . \\
    a_{n1} & a_{n2} & . & . & . & a_{nm} \\
  \end{array}
\right]^{T} \\&=&\left[
  \begin{array}{cccccc}
    b_{11} & b_{12} & . & . & . & b_{1n} \\
    b_{21} & b_{22} & . & . & . & b_{2n} \\
    . & . & . & . & . & . \\
    . & . & . & . & . & . \\
    . & . & . & . & . & . \\
    b_{n1} & b_{n2} & . & . & . & b_{nn} \\
  \end{array}
\right];\end{eqnarray*}
\begin{eqnarray*}
\Gamma(\mathscr{C}^{+})&=&M_{\mathscr{C}^{+}}\cdot
M_{\mathscr{C}^{+}}^{T}\\&=&\left[
  \begin{array}{ccccccccccc}
    a_{11} & a_{12} & . & . & . & a_{1m}& a_{1(m+1)}& . & . & . & a_{1(m+l)} \\
    a_{21} & a_{22} & . & . & . & a_{2m}& a_{2(m+1)} & . & . & . & a_{2(m+l)}\\
    . & . & . & . & . & . & . & . & . & . & .\\
    . & . & . & . & . & . & . & . & . & . & .\\
    . & . & . & . & . & . & . & . & . & . & .\\
    a_{n1} & a_{n2} & . & . & . & a_{nm}& a_{n(m+1)} & . & . & . & a_{n(m+l)}\\
    a_{(n+1)1} & a_{(n+1)2} & . & . & . & a_{(n+1)m}& a_{(n+1)(m+1)} & . & . & . & a_{(n+1)(m+l)}\\
    . & . & . & . & . & .& . & . & . & . & .\\
    . & . & . & . & . & .&. & . & . & . & .\\
    . & . & . & . & . & .& . & . & . & . & .\\
    a_{(n+t)1} & a_{(n+t)2} & . & . & . & a_{(n+t)m}& a_{(n+t)(m+1)} & . & . & . & a_{(n+t)(m+l)}\\
  \end{array}
\right]\\&&\cdot \left[
  \begin{array}{ccccccccccc}
    a_{11} & a_{12} & . & . & . & a_{1m}& a_{1(m+1)}& . & . & . & a_{1(m+l)} \\
    a_{21} & a_{22} & . & . & . & a_{2m}& a_{2(m+1)} & . & . & . & a_{2(m+l)}\\
    . & . & . & . & . & . & . & . & . & . & .\\
    . & . & . & . & . & . & . & . & . & . & .\\
    . & . & . & . & . & . & . & . & . & . & .\\
    a_{n1} & a_{n2} & . & . & . & a_{nm}& a_{n(m+1)} & . & . & . & a_{n(m+l)}\\
    a_{(n+1)1} & a_{(n+1)2} & . & . & . & a_{(n+1)m}& a_{(n+1)(m+1)} & . & . & . & a_{(n+1)(m+l)}\\
    . & . & . & . & . & .& . & . & . & . & .\\
    . & . & . & . & . & .&. & . & . & . & .\\
    . & . & . & . & . & .& . & . & . & . & .\\
    a_{(n+t)1} & a_{(n+t)2} & . & . & . & a_{(n+t)m}& a_{(n+t)(m+1)} & . & . & . & a_{(n+t)(m+l)}\\
  \end{array}
\right]^{T} \\
&=&\left[
  \begin{array}{ccccccccccccccc}
    c_{11} & c_{12} & . & . & . & c_{1n}& c_{1(n+1)}& . & . & . &c_{1(n+t)}\\
    c_{21} & c_{22} & . & . & . & c_{2n}& c_{2(n+1)}& . & . & . &c_{2(n+t)}\\
    . & . & . & . & . & . & . & . & . & . &.\\
    . & . & . & . & . & . & . & . & . & . &.\\
    . & . & . & . & . & . & . & . & . & . &.\\
    c_{n1} & c_{n2} & . & . & .& c_{nn}& c_{n(n+1)} &. & . & . &c_{n(n+t)}\\
    c_{(n+1)1} & c_{(n+1)2} & . & . & . & c_{(n+1)n}& c_{(n+1)(n+1)}& . & . & . &c_{(n+1)(n+t)}\\
    . & . & . & . & . & .& .& . & . & . &.\\
    . & . & . & . & . & .& .& . & . & . &.\\
    . & . & . & . & . & .& .& . & . & . &.\\
    c_{(n+t)1} & c_{(n+t)2} & . & . & . & c_{(n+t)n}& c_{(n+t)(n+1)}& . & . & . &c_{(n+t)(n+t)}\\
  \end{array}
\right];
\end{eqnarray*}

In the sense of the type-1 characteristic matrice of
$\mathscr{C}^{+}$, we have
\begin{eqnarray*}
c_{11}&=&\left[
  \begin{array}{ccccccccccc}
    a_{11} & a_{12} & . & . & . & a_{1m}& a_{1(m+1)}& . & . & . & a_{1(m+l)} \\
  \end{array}
\right]\\&&\cdot \left[
  \begin{array}{ccccccccccc}
    a_{11} & a_{12} & . & . & . & a_{1m}& a_{1(m+1)}& . & . & . & a_{1(m+l)} \\
  \end{array}
\right]^{T} \\
&=&\left[
  \begin{array}{ccccccccccc}
    a_{11} & a_{12} & . & . & . & a_{1m}\\
  \end{array}
\right]\cdot \left[
  \begin{array}{ccccccccccc}
    a_{11} & a_{12} & . & . & . & a_{1m}\\
  \end{array}
\right]^{T}\\&&\vee\left[
  \begin{array}{ccccc}
     a_{1(m+1)}& . & . & . & a_{1(m+l)} \\
  \end{array}
\right]\cdot \left[
  \begin{array}{ccccc}
     a_{1(m+1)}& . & . & . & a_{1(m+l)} \\
  \end{array}
\right]^{T}\\&=&b_{11}\vee\left[
  \begin{array}{ccccc}
     a_{1(m+1)}& . & . & . & a_{1(m+l)} \\
  \end{array}
\right]\cdot \left[
  \begin{array}{ccccc}
     a_{1(m+1)}& . & . & . & a_{1(m+l)} \\
  \end{array}
\right]^{T};\\
c_{(n+1)1}&=&\left[
  \begin{array}{ccccccccccc}
    a_{(n+1)1} & a_{(n+1)2} & . & . & . & a_{(n+1)m}& a_{(n+1)(m+1)}& . & . & . & a_{(n+1)(m+l)} \\
  \end{array}
\right]\\&&\cdot \left[
  \begin{array}{ccccccccccc}
    a_{11} & a_{12} & . & . & . & a_{1m}& a_{1(m+1)}& . & . & . & a_{1(m+l)} \\
  \end{array}
\right]^{T}\\&=&0\vee \left[
  \begin{array}{ccccccccccc}
    a_{11} & a_{12} & . & . & . & a_{1m}& a_{1(m+1)}& . & . & . & a_{1(m+l)} \\
  \end{array}
\right]^{T};
\end{eqnarray*}
\begin{eqnarray*}
c_{(n+1)(n+1)}&=&\left[
  \begin{array}{ccccccccccc}
    a_{(n+1)1} & a_{(n+1)2} & . & . & . & a_{(n+1)m}& a_{(n+1)(m+1)}& . & . & . & a_{(n+1)(m+l)} \\
  \end{array}
\right]\\&&\cdot \left[
  \begin{array}{ccccccccccc}
    a_{(n+1)1} & a_{(n+1)2} & . & . & . & a_{(n+1)m}& a_{(n+1)(m+1)}& . & . & . & a_{(n+1)(m+l)} \\
  \end{array}
\right]^{T}\\&=&0\vee\left[
  \begin{array}{ccccccccccc}
    a_{(n+1)1} & a_{(n+1)2} & . & . & . & a_{(n+1)m}& a_{(n+1)(m+1)}& . & . & . & a_{(n+1)(m+l)} \\
  \end{array}
\right]\\&&\cdot \left[
  \begin{array}{ccccccccccc}
    a_{(n+1)1} & a_{(n+1)2} & . & . & . & a_{(n+1)m}& a_{(n+1)(m+1)}& . & . & . & a_{(n+1)(m+l)} \\
  \end{array}
\right]^{T}.
\end{eqnarray*}

Since
$c_{11}\in\triangle_{1}(\Gamma(\mathscr{C}))$,
$c_{(n+1)1}\in\triangle_{2}(\Gamma(\mathscr{C}))$ and $c_{(n+1)(n+1)}\in\triangle_{3}(\Gamma(\mathscr{C}))$, we can compute other elements of $\triangle_{1}(\Gamma(\mathscr{C}))$,
$\triangle_{2}(\Gamma(\mathscr{C}))$ and $\triangle_{3}(\Gamma(\mathscr{C}))$ similarly.
Thus,
to obtain $\Gamma(\mathscr{C}^{+})$, we only need to compute
$\triangle_{1}(\Gamma(\mathscr{C}))$,
$\triangle_{2}(\Gamma(\mathscr{C}))$ and $\triangle_{3}(\Gamma(\mathscr{C}))$
on the basis of
$\Gamma(\mathscr{C})$ as follows:
  \begin{eqnarray*}
  \triangle_{1}(\Gamma(\mathscr{C}))&=&\left[\begin{array}{cccccc}
  a_{1(m+1)}&a_{2(m+1)}&.&.&. & a_{n(m+1)}\\
  a_{1(m+2)}&a_{2(m+2)}&.&.&. & a_{n(m+2)}\\
  .&.&.&.&. & .\\
  .&.&.&.&. & .\\
  .&.&.&.&. & .\\
  a_{1(m+l)}&a_{2(m+l)}&.&.&. & a_{n(m+l)}\\
  \end{array}
\right]^{T}\cdot \left[\begin{array}{cccccc}
  a_{1(m+1)}&a_{2(m+1)}&.&.&. & a_{n(m+1)}\\
  a_{1(m+2)}&a_{2(m+2)}&.&.&. & a_{n(m+2)}\\
  .&.&.&.&. & .\\
  .&.&.&.&. & .\\
  .&.&.&.&. & .\\
  a_{1(m+l)}&a_{2(m+l)}&.&.&. & a_{n(m+l)}\\
  \end{array}
\right];
 \\
\triangle_{2}(\Gamma(\mathscr{C}))&=&\left[\begin{array}{cccccc}
  a_{(n+1)1}&a_{(n+1)2}&.&.&. & a_{(n+1)(m+l)}\\
  a_{(n+2)1}&a_{(n+2)2}&.&.&. & a_{(n+2)(m+l)}\\
  .&.&.&.&. & .\\
  .&.&.&.&. & .\\
  .&.&.&.&. & .\\
  a_{(n+t)1}&a_{(n+t)2}&.&.&. & a_{(n+t)(m+l)}\\
  \end{array}
\right]\cdot \left[\begin{array}{cccccc}
  a_{11}&a_{12}&.&.&. & a_{1(m+l)} \\
  a_{21}&a_{22}&.&.&. & a_{2(m+l)} \\
  .&.&.&.&. & . \\
  .&.&.&.&. & . \\
  .&.&.&.&. & . \\
  a_{n1}&a_{n2}&.&.&. & a_{n(m+l)} \\
  \end{array}
\right]^{T};
\\
 \triangle_{3}(\Gamma(\mathscr{C}))&=&\left[\begin{array}{cccccc}
  a_{(n+1)1}&a_{(n+1)2}&.&.&. & a_{(n+1)(m+l)}\\
  a_{(n+2)1}&a_{(n+2)2}&.&.&. & a_{(n+2)(m+l)}\\
  .&.&.&.&. & .\\
  .&.&.&.&. & .\\
  .&.&.&.&. &.\\
  a_{(n+t)1}&a_{(n+t)2}&.&.&. & a_{(n+t)(m+l)}\\
  \end{array}
\right]\cdot \left[\begin{array}{cccccc}
  a_{(n+1)1}&a_{(n+1)2}&.&.&. & a_{(n+1)(m+l)}\\
  a_{(n+2)1}&a_{(n+2)2}&.&.&. & a_{(n+2)(m+l)}\\
  .&.&.&.&. & .\\
  .&.&.&.&. & .\\
  .&.&.&.&. &.\\
  a_{(n+t)1}&a_{(n+t)2}&.&.&. & a_{(n+t)(m+l)}\\
  \end{array}
\right]^{T}.
\end{eqnarray*}

Therefore, we have
\begin{eqnarray*}
 \Gamma(\mathscr{C}^{+})=\left[
  \begin{array}{cc}
    \Gamma(\mathscr{C}) & 0\\
    0 & 0\\
  \end{array}
\right]\bigvee \left[
  \begin{array}{cc}
    \triangle_{1}(\Gamma(\mathscr{C})) & (\triangle_{2}(\Gamma(\mathscr{C})))^{T}\\
    \triangle_{2}(\Gamma(\mathscr{C})) & \triangle_{3}(\Gamma(\mathscr{C}))\\
  \end{array}
\right].
\end{eqnarray*}

\begin{example}
Let $U=\{x_{1},x_{2},x_{3},x_{4}\}$, $U^{+}=U\cup\{x_{5},x_{6}\}$,
$\mathscr{C}=\{C_{1},C_{2},C_{3}\}$,
$\mathscr{C}^{+}=\{C^{+}_{1},C^{+}_{2},C^{+}_{3},C^{+}_{4},C^{+}_{5}\}$, where
$C_{1}=\{x_{1},x_{4}\}$, $C_{2}=\{x_{1},x_{2},x_{4}\}$,
$C_{3}=\{x_{3},x_{4}\}$, $C^{+}_{1}=\{x_{1},x_{4},x_{5}\}$,
$C^{+}_{2}=\{x_{1},x_{2},x_{4},x_{5}\}$,
$C^{+}_{3}=\{x_{3},x_{4}\}$, $C^{+}_{4}=\{x_{3},x_{5},x_{6}\}$,$C^{+}_{5}=\{x_{1},x_{6}\}$, and
$X=\{x_{3},x_{4},x_{5}\}$. By Definition 3.1, we first have that
\begin{eqnarray*}
\Gamma(\mathscr{C})&=&M_{\mathscr{C}}\cdot
M_{\mathscr{C}}^{T}=\left[
\begin{array}{cccc}
1 & 1 & 0 \\
0 & 1 & 0 \\
0 & 0 & 1 \\
1 & 1 & 1 \\
\end{array}
\right] \cdot \left[
\begin{array}{cccc}
1 & 1 & 0 \\
0 & 1 & 0 \\
0 & 0 & 1 \\
1 & 1 & 1 \\
\end{array}
\right]^{T}  =\left[
\begin{array}{cccc}
1 & 1 & 0 & 1 \\
1 & 1 & 0 & 1 \\
0 & 0 & 1 & 1 \\
1 & 1 & 1 & 1 \\
\end{array}
\right].
\end{eqnarray*}

Second, by Theorem 3.2, we get that
\begin{eqnarray*}
\triangle_{1}(\Gamma(\mathscr{C}))&=&\left[
\begin{array}{ccccc}
0 & 0 & 1 & 0 \\
1 & 0 & 0 & 0 \\
\end{array}
\right]^{T}\cdot \left[
\begin{array}{cccccc}
0 & 0 & 1 & 0 \\
1 & 0 & 0 & 0 \\
\end{array}
\right]=\left[
\begin{array}{ccccc}
1 & 0 & 0 & 0  \\
0 & 0 & 0 & 0  \\
0 & 0 & 1 & 0  \\
0 & 0 & 0 & 0  \\
\end{array}
\right];
\end{eqnarray*}
\begin{eqnarray*} \triangle_{2}(\Gamma(\mathscr{C}))&=&\left[
\begin{array}{ccccc}
1 & 1 & 0 & 1& 0\\
0 & 0 & 1 & 0& 1\\
\end{array}
\right]\cdot \left[
\begin{array}{cccccc}
1 & 1 & 0 & 0 & 1\\
0 & 1 & 0 & 0 & 0\\
0 & 0 & 1 & 1 & 0\\
1 & 1 & 1 & 0 & 0\\
\end{array}
\right]^{T}=\left[
\begin{array}{cccc}
1 & 1 & 1 & 1 \\
1 & 0 & 1 & 1 \\
\end{array}
\right];\\
\triangle_{3}(\Gamma(\mathscr{C}))&=&\left[
\begin{array}{ccccc}
1 & 1 & 0 & 1 & 0\\
0 & 0 & 1 & 0 & 1\\
\end{array}
\right]\cdot \left[
\begin{array}{ccccc}
1 & 1 & 0 & 1 & 0\\
0 & 0 & 1 & 0 & 1\\
\end{array}
\right]^{T}=\left[
\begin{array}{cc}
1 & 0 \\
0 & 1 \\
\end{array}
\right].
\end{eqnarray*}

Thus, we obtain that
\begin{eqnarray*}
 \Gamma(\mathscr{C}^{+})&=&(c_{ij})_{66}\\&=&\left[
  \begin{array}{cc}
    \Gamma(\mathscr{C}) & 0\\
    0 & 0\\
  \end{array}
\right]\bigvee \left[
  \begin{array}{cc}
    \triangle_{1}(\Gamma(\mathscr{C}) & (\triangle_{2}(\Gamma(\mathscr{C})))^{T}\\
    \triangle_{2}(\Gamma(\mathscr{C})) & \triangle_{3}(\Gamma(\mathscr{C}))\\
  \end{array}
\right]\\ &=&\left[
\begin{array}{cccccc}
1 & 1 & 0 & 1 & 0& 0\\
1 & 1 & 0 & 1 & 0& 0\\
0 & 0 & 1 & 1 & 0& 0\\
1 & 1 & 1 & 1 & 0& 0\\
0 & 0 & 0 & 0 & 0& 0\\
0 & 0 & 0 & 0 & 0& 0\\
\end{array}
\right]\bigvee \left[
\begin{array}{cccccc}
1 & 0 & 0 & 0 & 1& 1\\
0 & 0 & 0 & 0 & 1& 0\\
0 & 0 & 1 & 0 & 1& 1\\
0 & 0 & 0 & 0 & 1& 1\\
1 & 1 & 1 & 1 & 1& 0\\
1 & 0 & 1 & 1 & 0& 1\\
\end{array}
\right]\\&=& \left[
\begin{array}{cccccc}
1 & 1 & 0 & 1 & 1& 1\\
1 & 1 & 0 & 1 & 1& 0\\
0 & 0 & 1 & 1 & 1& 1\\
1 & 1 & 1 & 1 & 1& 1\\
1 & 1 & 1 & 1 & 1& 0\\
1 & 0 & 1 & 1 & 0& 1\\
\end{array}
\right].
\end{eqnarray*}
By Definition 2.5, we have that
\begin{eqnarray*}
\mathcal {X}_{SH(X)}&=&\Gamma(\mathscr{C}^{+})\cdot \mathcal {X}_{X}
=\left[
\begin{array}{cccccc}
1 & 1 & 0 & 1 & 1& 1\\
1 & 1 & 0 & 1 & 1& 0\\
0 & 0 & 1 & 1 & 1& 1\\
1 & 1 & 1 & 1 & 1& 1\\
1 & 1 & 1 & 1 & 1& 0\\
1 & 0 & 1 & 1 & 0& 1\\
\end{array}
\right]\cdot\left[
\begin{array}{c}
0 \\
0 \\
1 \\
1 \\
1 \\
0 \\
\end{array}
\right] =\left[
\begin{array}{cccccc}
1 & 1 & 1 & 1 & 1 & 1\\
\end{array}
\right]^{T};
\\
 \mathcal {X}_{SL(X)}&=&\Gamma(\mathscr{C}^{+})\odot
\mathcal {X}_{X}=\left[
\begin{array}{cccccc}
1 & 1 & 0 & 1 & 1& 1\\
1 & 1 & 0 & 1 & 1& 0\\
0 & 0 & 1 & 1 & 1& 1\\
1 & 1 & 1 & 1 & 1& 1\\
1 & 1 & 1 & 1 & 1& 0\\
1 & 0 & 1 & 1 & 0& 1\\
\end{array}
\right]\odot \left[
\begin{array}{c}
0 \\
0 \\
1 \\
1 \\
1 \\
0 \\
\end{array}
\right]=\left[
\begin{array}{ccccccc}
0 & 0 & 1 & 0 & 0& 0\\
\end{array}
\right]^{T}.
\end{eqnarray*}

Therefore, $SH(X)=\{x_{1},x_{2},x_{3},x_{4},x_{5},x_{6}\}$ and
$SL(X)=\{x_{3}\}$.
\end{example}

In Example 3.3, we only need to calculate elements in
$\triangle_{1}(\Gamma(\mathscr{C}))$,
$\triangle_{2}(\Gamma(\mathscr{C}))$ and $\triangle_{3}(\Gamma(\mathscr{C}))$ by Theorem 3.2.
Thereby, the incremental algorithm is effective to compute the
second lower and upper approximations of sets.

In practical situations, there exists a need to construct the type-2
characteristic matrices of dynamic coverings for computing the sixth
lower and upper approximations of sets. Subsequently, we construct
$\prod(\mathscr{C}^{+})$ based on $\prod(\mathscr{C})$. For
convenience, we denote $\prod(\mathscr{C})=(d_{ij})_{n\times n}$ and
$\prod(\mathscr{C}^{+})=(e_{ij})_{(n+t)\times (n+t)}$.

\begin{theorem}
Let $(U^{+},\mathscr{C}^{+})$ be a dynamic covering approximation space
of $(U,\mathscr{C})$, $\prod(\mathscr{C})$ and
$\prod(\mathscr{C}^{+})$ the type-2 characteristic matrices of
$\mathscr{C}$ and $\mathscr{C}^{+}$, respectively. Then
\begin{eqnarray*}
 \prod(\mathscr{C}^{+})=\left[
  \begin{array}{cc}
   \prod(\mathscr{C}) & 1\\
    1 & 1\\
  \end{array}
\right]\bigwedge \left[
  \begin{array}{cc}
    \triangle_{1}(\prod(\mathscr{C})) & \triangle_{3}(\prod(\mathscr{C}))\\
    \triangle_{2}(\prod(\mathscr{C})) &\triangle_{4}(\prod(\mathscr{C}))\\
  \end{array}
\right], \end{eqnarray*}
where
  \begin{eqnarray*}
  \triangle_{1}(\prod(\mathscr{C}))&=&\left[\begin{array}{cccccc}
  a_{1(m+1)}&a_{2(m+1)}&.&.&. & a_{n(m+1)}\\
  a_{1(m+2)}&a_{2(m+2)}&.&.&. & a_{n(m+2)}\\
  .&.&.&.&. & .\\
  .&.&.&.&. & .\\
  .&.&.&.&. & .\\
  a_{1(m+l)}&a_{2(m+l)}&.&.&. & a_{n(m+l)}\\
  \end{array}
\right]^{T}\odot \left[\begin{array}{cccccc}
  a_{1(m+1)}&a_{2(m+1)}&.&.&. & a_{n(m+1)}\\
  a_{1(m+2)}&a_{2(m+2)}&.&.&. & a_{n(m+2)}\\
  .&.&.&.&. & .\\
  .&.&.&.&. & .\\
  .&.&.&.&. & .\\
  a_{1(m+l)}&a_{2(m+l)}&.&.&. & a_{n(m+l)}\\
  \end{array}
\right];
 \\
\triangle_{2}(\prod(\mathscr{C}))&=&\left[\begin{array}{cccccc}
  a_{(n+1)1}&a_{(n+1)2}&.&.&. & a_{(n+1)(m+l)}\\
  a_{(n+2)1}&a_{(n+2)2}&.&.&. & a_{(n+2)(m+l)}\\
  .&.&.&.&. & .\\
  .&.&.&.&. & .\\
  .&.&.&.&. & .\\
  a_{(n+t)1}&a_{(n+t)2}&.&.&. & a_{(n+t)(m+l)}\\
  \end{array}
\right]\odot \left[\begin{array}{cccccc}
  a_{11}&a_{12}&.&.&. & a_{1(m+l)} \\
  a_{21}&a_{22}&.&.&. & a_{2(m+l)} \\
  .&.&.&.&. & . \\
  .&.&.&.&. & . \\
  .&.&.&.&. & . \\
  a_{n1}&a_{n2}&.&.&. & a_{n(m+l)} \\
  \end{array}
\right]^{T};\\
\\
\triangle_{3}(\prod(\mathscr{C}))&=&\left[\begin{array}{cccccc}
  a_{11}&a_{12}&.&.&. & a_{1(m+l)}\\
  a_{21}&a_{22}&.&.&. & a_{2(m+l)}\\
  .&.&.&.&. & .\\
  .&.&.&.&. & .\\
  .&.&.&.&. & .\\
  a_{n1}&a_{n2}&.&.&. & a_{n(m+l)}\\
  \end{array}
\right]\odot \left[\begin{array}{cccccc}
  a_{(n+1)1}&a_{(n+1)2}&.&.&. & a_{(n+1)(m+l)}\\
  a_{(n+2)1}&a_{(n+2)2}&.&.&. & a_{(n+2)(m+l)}\\
  .&.&.&.&. & .\\
  .&.&.&.&. & .\\
  .&.&.&.&. & .\\
  a_{(n+t)1}&a_{(n+t)2}&.&.&. & a_{(n+t)(m+l)}\\
  \end{array}
\right]^{T};
\\
 \triangle_{4}(\prod(\mathscr{C}))&=&\left[\begin{array}{cccccc}
  a_{(n+1)1}&a_{(n+1)2}&.&.&. & a_{(n+1)(m+l)}\\
  a_{(n+2)1}&a_{(n+2)2}&.&.&. & a_{(n+2)(m+l)}\\
  .&.&.&.&. & .\\
  .&.&.&.&. & .\\
  .&.&.&.&. &.\\
  a_{(n+t)1}&a_{(n+t)2}&.&.&. & a_{(n+t)(m+l)}\\
  \end{array}
\right]\odot \left[\begin{array}{cccccc}
  a_{(n+1)1}&a_{(n+1)2}&.&.&. & a_{(n+1)(m+l)}\\
  a_{(n+2)1}&a_{(n+2)2}&.&.&. & a_{(n+2)(m+l)}\\
  .&.&.&.&. & .\\
  .&.&.&.&. & .\\
  .&.&.&.&. &.\\
  a_{(n+t)1}&a_{(n+t)2}&.&.&. & a_{(n+t)(m+l)}\\
  \end{array}
\right]^{T}.
\end{eqnarray*}
\end{theorem}

\noindent\textbf{Proof.} By Definition 3.1, we have
$\prod(\mathscr{C})$ and $\prod(\mathscr{C}^{+})$ as follows:
\begin{eqnarray*}
\prod(\mathscr{C})&=&M_{\mathscr{C}}\odot M_{\mathscr{C}}^{T}\\&=&\left[
  \begin{array}{cccccc}
    a_{11} & a_{12} & . & . & . & a_{1m} \\
    a_{21} & a_{22} & . & . & . & a_{2m} \\
    . & . & . & . & . & . \\
    . & . & . & . & . & . \\
    . & . & . & . & . & . \\
    a_{n1} & a_{n2} & . & . & . & a_{nm} \\
  \end{array}
\right] \odot \left[
  \begin{array}{cccccc}
    a_{11} & a_{12} & . & . & . & a_{1m} \\
    a_{21} & a_{22} & . & . & . & a_{2m} \\
    . & . & . & . & . & . \\
    . & . & . & . & . & . \\
    . & . & . & . & . & . \\
    a_{n1} & a_{n2} & . & . & . & a_{nm} \\
  \end{array}
\right]^{T} \\&=&\left[
  \begin{array}{cccccc}
    d_{11} & d_{12} & . & . & . & d_{1n} \\
    d_{21} & d_{22} & . & . & . & d_{2n} \\
    . & . & . & . & . & . \\
    . & . & . & . & . & . \\
    . & . & . & . & . & . \\
    d_{n1} & d_{n2} & . & . & . & d_{nn} \\
  \end{array}
\right];
\end{eqnarray*}\begin{eqnarray*}
\prod(\mathscr{C}^{+})&=&M_{\mathscr{C}^{+}}\odot
M_{\mathscr{C}^{+}}^{T}\\&=&\left[
  \begin{array}{ccccccccccc}
    a_{11} & a_{12} & . & . & . & a_{1m}& a_{1(m+1)}& . & . & . & a_{1(m+l)} \\
    a_{21} & a_{22} & . & . & . & a_{2m}& a_{2(m+1)} & . & . & . & a_{2(m+l)}\\
    . & . & . & . & . & . & . & . & . & . & .\\
    . & . & . & . & . & . & . & . & . & . & .\\
    . & . & . & . & . & . & . & . & . & . & .\\
    a_{n1} & a_{n2} & . & . & . & a_{nm}& a_{n(m+1)} & . & . & . & a_{n(m+l)}\\
    a_{(n+1)1} & a_{(n+1)2} & . & . & . & a_{(n+1)m}& a_{(n+1)(m+1)} & . & . & . & a_{(n+1)(m+l)}\\
    . & . & . & . & . & .& . & . & . & . & .\\
    . & . & . & . & . & .&. & . & . & . & .\\
    . & . & . & . & . & .& . & . & . & . & .\\
    a_{(n+t)1} & a_{(n+t)2} & . & . & . & a_{(n+t)m}& a_{(n+t)(m+1)} & . & . & . & a_{(n+t)(m+l)}\\
  \end{array}
\right]\\&&\odot \left[
  \begin{array}{ccccccccccc}
    a_{11} & a_{12} & . & . & . & a_{1m}& a_{1(m+1)}& . & . & . & a_{1(m+l)} \\
    a_{21} & a_{22} & . & . & . & a_{2m}& a_{2(m+1)} & . & . & . & a_{2(m+l)}\\
    . & . & . & . & . & . & . & . & . & . & .\\
    . & . & . & . & . & . & . & . & . & . & .\\
    . & . & . & . & . & . & . & . & . & . & .\\
    a_{n1} & a_{n2} & . & . & . & a_{nm}& a_{n(m+1)} & . & . & . & a_{n(m+l)}\\
    a_{(n+1)1} & a_{(n+1)2} & . & . & . & a_{(n+1)m}& a_{(n+1)(m+1)} & . & . & . & a_{(n+1)(m+l)}\\
    . & . & . & . & . & .& . & . & . & . & .\\
    . & . & . & . & . & .&. & . & . & . & .\\
    . & . & . & . & . & .& . & . & . & . & .\\
    a_{(n+t)1} & a_{(n+t)2} & . & . & . & a_{(n+t)m}& a_{(n+t)(m+1)} & . & . & . & a_{(n+t)(m+l)}\\
  \end{array}
\right]^{T} \\ \end{eqnarray*} \begin{eqnarray*}
&=&\left[
  \begin{array}{ccccccccccccccc}
    e_{11} & e_{12} & . & . & . & e_{1n}& e_{1(n+1)}& . & . & . &e_{1(n+t)}\\
    e_{21} & e_{22} & . & . & . & e_{2n}& e_{2(n+1)}& . & . & . &e_{1(n+t)}\\
    . & . & . & . & . & . & . & . & . & . &.\\
    . & . & . & . & . & . & . & . & . & . &.\\
    . & . & . & . & . & . & . & . & . & . &.\\
    e_{n1} & e_{n2} & . & . & .& e_{nn}& e_{n(n+1)} &. & . & . &e_{1(n+t)}\\
    e_{(n+1)1} & e_{(n+1)2} & . & . & . & e_{(n+1)n}& e_{(n+1)(n+1)}& . & . & . &e_{1(n+1)}\\
    . & . & . & . & . & .& .& . & . & . &.\\
    . & . & . & . & . & .& .& . & . & . &.\\
    . & . & . & . & . & .& .& . & . & . &.\\
    e_{(n+t)1} & e_{(n+t)2} & . & . & . & e_{(n+t)n}& e_{(n+t)(n+1)}& . & . & . &e_{(n+t)(n+t)}\\
  \end{array}
\right].
\end{eqnarray*}

In the sense of the type-2 characteristic matrice of
$\mathscr{C}^{+}$, we have
\begin{eqnarray*}
e_{11}&=&\left[
  \begin{array}{ccccccccccc}
    a_{11} & a_{12} & . & . & . & a_{1m}& a_{1(m+1)}& . & . & . & a_{1(m+l)} \\
  \end{array}
\right]\\&&\odot \left[
  \begin{array}{ccccccccccc}
    a_{11} & a_{12} & . & . & . & a_{1m}& a_{1(m+1)}& . & . & . & a_{1(m+l)} \\
  \end{array}
\right]^{T} \\
&=&\left[
  \begin{array}{ccccccccccc}
    a_{11} & a_{12} & . & . & . & a_{1m}\\
  \end{array}
\right]\odot \left[
  \begin{array}{ccccccccccc}
    a_{11} & a_{12} & . & . & . & a_{1m}\\
  \end{array}
\right]^{T}\\&&\wedge\left[
  \begin{array}{ccccc}
     a_{1(m+1)}& . & . & . & a_{1(m+l)} \\
  \end{array}
\right]\odot \left[
  \begin{array}{ccccc}
     a_{1(m+1)}& . & . & . & a_{1(m+l)} \\
  \end{array}
\right]^{T}\\&=&d_{11}\wedge\left[
  \begin{array}{ccccc}
     a_{1(m+1)}& . & . & . & a_{1(m+l)} \\
  \end{array}
\right]\odot \left[
  \begin{array}{ccccc}
     a_{1(m+1)}& . & . & . & a_{1(m+l)} \\
  \end{array}
\right]^{T};\\
e_{(n+1)1}&=&\left[
  \begin{array}{ccccccccccc}
    a_{(n+1)1} & a_{(n+1)2} & . & . & . & a_{(n+1)m}& a_{(n+1)(m+1)}& . & . & . & a_{(n+1)(m+l)} \\
  \end{array}
\right]\\&&\odot \left[
  \begin{array}{ccccccccccc}
    a_{11} & a_{12} & . & . & . & a_{1m}& a_{1(m+1)}& . & . & . & a_{1(m+l)} \\
  \end{array}
\right]^{T}\\&=&1\wedge\left[
  \begin{array}{ccccccccccc}
    a_{(n+1)1} & a_{(n+1)2} & . & . & . & a_{(n+1)m}& a_{(n+1)(m+1)}& . & . & . & a_{(n+1)(m+l)} \\
  \end{array}
\right]\\&&\odot \left[
  \begin{array}{ccccccccccc}
    a_{11} & a_{12} & . & . & . & a_{1m}& a_{1(m+1)}& . & . & . & a_{1(m+l)} \\
  \end{array}
\right]^{T};\end{eqnarray*}\begin{eqnarray*}
e_{1(n+1)}&=&\left[
  \begin{array}{ccccccccccc}
    a_{11} & a_{12} & . & . & . & a_{1m}& a_{1(m+1)}& . & . & . & a_{1(m+l)} \\
  \end{array}
\right]\\&&\odot \left[
  \begin{array}{ccccccccccc}
    a_{(n+1)1} & a_{(n+1)2} & . & . & . & a_{(n+1)m}& a_{(n+1)(m+1)}& . & . & . & a_{(n+1)(m+l)} \\
  \end{array}
\right]^{T}\\&=&1\wedge\left[
  \begin{array}{ccccccccccc}
    a_{11} & a_{12} & . & . & . & a_{1m}& a_{1(m+1)}& . & . & . & a_{1(m+l)} \\
  \end{array}
\right]\\&&\odot \left[
  \begin{array}{ccccccccccc}
    a_{(n+1)1} & a_{(n+1)2} & . & . & . & a_{(n+1)m}& a_{(n+1)(m+1)}& . & . & . & a_{(n+1)(m+l)} \\
  \end{array}
\right]^{T};\\
e_{(n+1)(n+1)}&=&\left[
  \begin{array}{ccccccccccc}
    a_{(n+1)1} & a_{(n+1)2} & . & . & . & a_{(n+1)m}& a_{(n+1)(m+1)}& . & . & . & a_{(n+1)(m+l)} \\
  \end{array}
\right]\\&&\odot \left[
  \begin{array}{ccccccccccc}
    a_{(n+1)1} & a_{(n+1)2} & . & . & . & a_{(n+1)m}& a_{(n+1)(m+1)}& . & . & . & a_{(n+1)(m+l)} \\
  \end{array}
\right]^{T}\\&=&1\wedge\left[
  \begin{array}{ccccccccccc}
    a_{(n+1)1} & a_{(n+1)2} & . & . & . & a_{(n+1)m}& a_{(n+1)(m+1)}& . & . & . & a_{(n+1)(m+l)} \\
  \end{array}
\right]\\&&\odot \left[
  \begin{array}{ccccccccccc}
    a_{(n+1)1} & a_{(n+1)2} & . & . & . & a_{(n+1)m}& a_{(n+1)(m+1)}& . & . & . & a_{(n+1)(m+l)} \\
  \end{array}
\right]^{T}.
\end{eqnarray*}

Since
$e_{11}\in\triangle_{1}(\prod(\mathscr{C}))$,
$e_{(n+1)1}\in\triangle_{2}(\prod(\mathscr{C}))$,
$e_{1(n+1)}\in\triangle_{3}(\prod(\mathscr{C}))$ and $e_{(n+1)(n+1)}\in\triangle_{3}(\prod(\mathscr{C}))$, we can compute other elements of $\triangle_{1}(\prod(\mathscr{C}))$,
$\triangle_{2}(\prod(\mathscr{C}))$, $\triangle_{3}(\prod(\mathscr{C}))$ and $\triangle_{4}(\prod(\mathscr{C}))$ similarly.
Thus,
to compute $\prod(\mathscr{C}^{+})$ on the basis of
$\prod(\mathscr{C})$, we only need to compute
$\triangle_{1}(\prod(\mathscr{C}))$,
$\triangle_{2}(\prod(\mathscr{C}))$, $\triangle_{3}(\prod(\mathscr{C}))$
and $\triangle_{4}(\prod(\mathscr{C}))$ as follows:
   \begin{eqnarray*}
  \triangle_{1}(\prod(\mathscr{C}))&=&\left[\begin{array}{cccccc}
  a_{1(m+1)}&a_{2(m+1)}&.&.&. & a_{n(m+1)}\\
  a_{1(m+2)}&a_{2(m+2)}&.&.&. & a_{n(m+2)}\\
  .&.&.&.&. & .\\
  .&.&.&.&. & .\\
  .&.&.&.&. & .\\
  a_{1(m+l)}&a_{2(m+l)}&.&.&. & a_{n(m+l)}\\
  \end{array}
\right]^{T}\odot \left[\begin{array}{cccccc}
  a_{1(m+1)}&a_{2(m+1)}&.&.&. & a_{n(m+1)}\\
  a_{1(m+2)}&a_{2(m+2)}&.&.&. & a_{n(m+2)}\\
  .&.&.&.&. & .\\
  .&.&.&.&. & .\\
  .&.&.&.&. & .\\
  a_{1(m+l)}&a_{2(m+l)}&.&.&. & a_{n(m+l)}\\
  \end{array}
\right];
 \\
\triangle_{2}(\prod(\mathscr{C}))&=&\left[\begin{array}{cccccc}
  a_{(n+1)1}&a_{(n+1)2}&.&.&. & a_{(n+1)(m+l)}\\
  a_{(n+2)1}&a_{(n+2)2}&.&.&. & a_{(n+2)(m+l)}\\
  .&.&.&.&. & .\\
  .&.&.&.&. & .\\
  .&.&.&.&. & .\\
  a_{(n+t)1}&a_{(n+t)2}&.&.&. & a_{(n+t)(m+l)}\\
  \end{array}
\right]\odot \left[\begin{array}{cccccc}
  a_{11}&a_{12}&.&.&. & a_{1(m+l)} \\
  a_{21}&a_{22}&.&.&. & a_{2(m+l)} \\
  .&.&.&.&. & . \\
  .&.&.&.&. & . \\
  .&.&.&.&. & . \\
  a_{n1}&a_{n2}&.&.&. & a_{n(m+l)} \\
  \end{array}
\right]^{T};\\
\\
\triangle_{3}(\prod(\mathscr{C}))&=&\left[\begin{array}{cccccc}
  a_{11}&a_{12}&.&.&. & a_{1(m+l)}\\
  a_{21}&a_{22}&.&.&. & a_{2(m+l)}\\
  .&.&.&.&. & .\\
  .&.&.&.&. & .\\
  .&.&.&.&. & .\\
  a_{n1}&a_{n2}&.&.&. & a_{n(m+l)}\\
  \end{array}
\right]\odot \left[\begin{array}{cccccc}
  a_{(n+1)1}&a_{(n+1)2}&.&.&. & a_{(n+1)(m+l)}\\
  a_{(n+2)1}&a_{(n+2)2}&.&.&. & a_{(n+2)(m+l)}\\
  .&.&.&.&. & .\\
  .&.&.&.&. & .\\
  .&.&.&.&. & .\\
  a_{(n+t)1}&a_{(n+t)2}&.&.&. & a_{(n+t)(m+l)}\\
  \end{array}
\right]^{T};\\
 \triangle_{4}(\prod(\mathscr{C}))&=&\left[\begin{array}{cccccc}
  a_{(n+1)1}&a_{(n+1)2}&.&.&. & a_{(n+1)(m+l)}\\
  a_{(n+2)1}&a_{(n+2)2}&.&.&. & a_{(n+2)(m+l)}\\
  .&.&.&.&. & .\\
  .&.&.&.&. & .\\
  .&.&.&.&. &.\\
  a_{(n+t)1}&a_{(n+t)2}&.&.&. & a_{(n+t)(m+l)}\\
  \end{array}
\right]\odot \left[\begin{array}{cccccc}
  a_{(n+1)1}&a_{(n+1)2}&.&.&. & a_{(n+1)(m+l)}\\
  a_{(n+2)1}&a_{(n+2)2}&.&.&. & a_{(n+2)(m+l)}\\
  .&.&.&.&. & .\\
  .&.&.&.&. & .\\
  .&.&.&.&. &.\\
  a_{(n+t)1}&a_{(n+t)2}&.&.&. & a_{(n+t)(m+l)}\\
  \end{array}
\right]^{T}.
\end{eqnarray*}

Therefore, we have
\begin{eqnarray*}
 \prod(\mathscr{C}^{+})=\left[
  \begin{array}{cc}
    \prod(\mathscr{C}) & 1\\
    1 & 1\\
  \end{array}
\right]\bigwedge \left[
  \begin{array}{cc}
    \triangle_{1}(\prod(\mathscr{C})) & \triangle_{3}(\prod(\mathscr{C}))\\
    \triangle_{2}(\prod(\mathscr{C})) & \triangle_{4}(\prod(\mathscr{C}))\\
  \end{array}
\right].
\end{eqnarray*}

The following example illustrates that how to compute the sixth
lower and upper approximations of set by using the incremental
algorithm.
\begin{example} (Continuation of Example 3.3)
We obtain that
\begin{eqnarray*}
\prod(\mathscr{C})&=&M_{\mathscr{C}}\odot M_{\mathscr{C}}^{T}
=\left[
\begin{array}{cccc}
1 & 1 & 0 \\
0 & 1 & 0 \\
0 & 0 & 1 \\
1 & 1 & 1 \\
\end{array}
\right] \odot \left[
\begin{array}{cccc}
1 & 1 & 0 \\
0 & 1 & 0 \\
0 & 0 & 1 \\
1 & 1 & 1 \\
\end{array}
\right]^{T}=\left[
\begin{array}{cccc}
1 & 0 & 0 & 1 \\
1 & 1 & 0 & 1 \\
0 & 0 & 1 & 1 \\
0 & 0 & 0 & 1 \\
\end{array}
\right].
\end{eqnarray*}

By Theorem 3.4, we have that
\begin{eqnarray*}
\triangle_{1}(\prod(\mathscr{C}))&=&\left[
\begin{array}{ccccc}
0 & 0 & 1 & 0 \\
1 & 0 & 0 & 0 \\
\end{array}
\right]^{T}\odot \left[
\begin{array}{cccccc}
0 & 0 & 1 & 0 \\
1 & 0 & 0 & 0 \\
\end{array}
\right]=\left[
\begin{array}{ccccc}
1 & 0 & 0 & 0  \\
1 & 1 & 1 & 1  \\
0 & 0 & 1 & 0  \\
1 & 1 & 1 & 1  \\
\end{array}
\right];\\ \triangle_{2}(\prod(\mathscr{C}))&=&\left[
\begin{array}{ccccc}
1 & 1 & 0 & 1& 0\\
0 & 0 & 1 & 0& 1\\
\end{array}
\right]\odot \left[
\begin{array}{cccccc}
1 & 1 & 0 & 0 & 1\\
0 & 1 & 0 & 0 & 0\\
0 & 0 & 1 & 1 & 0\\
1 & 1 & 1 & 0 & 0\\
\end{array}
\right]^{T}=\left[
\begin{array}{cccc}
0 & 0 & 0 & 0 \\
0 & 0 & 0 & 0 \\
\end{array}
\right];\\
\triangle_{3}(\prod(\mathscr{C}))&=&\left[
\begin{array}{cccccc}
1 & 1 & 0 & 0 & 1\\
0 & 1 & 0 & 0 & 0\\
0 & 0 & 1 & 1 & 0\\
1 & 1 & 1 & 0 & 0\\
\end{array}
\right]^{T}\odot\left[
\begin{array}{ccccc}
1 & 1 & 0 & 1& 0\\
0 & 0 & 1 & 0& 1\\
\end{array}
\right] =\left[
\begin{array}{cccc}
0 & 1 & 0 & 0 \\
0 & 0 & 0 & 0 \\
\end{array}
\right];\\
\triangle_{4}(\prod(\mathscr{C}))&=&\left[
\begin{array}{ccccc}
1 & 1 & 0 & 1 & 0\\
0 & 0 & 1 & 0 & 1\\
\end{array}
\right]\odot \left[
\begin{array}{ccccc}
1 & 1 & 0 & 1 & 0\\
0 & 0 & 1 & 0 & 1\\
\end{array}
\right]^{T}=\left[
\begin{array}{cc}
1 & 0 \\
0 & 1 \\
\end{array}
\right].
\end{eqnarray*}
\end{example}

Thus, we have
\begin{eqnarray*}
\prod(\mathscr{C}^{+})&=&\left[
  \begin{array}{cc}
    \prod(\mathscr{C}) & 1\\
    1 & 1\\
  \end{array}
\right]\bigwedge\left[
  \begin{array}{cc}
    \triangle_{1}(\prod(\mathscr{C})) & \triangle_{3}(\prod(\mathscr{C}))\\
    \triangle_{2}(\prod(\mathscr{C})) & \triangle_{4}(\prod(\mathscr{C}))\\
  \end{array}
\right]=\left[
\begin{array}{ccccccc}
1  &   0   &  0   &  0  &   0&   0\\
1  &   1   &  0   &  1  &   1&   0\\
0  &   0   &  1   &  0  &   0&   0\\
0  &   0   &  0   &  1  &   0&   0\\
0  &   0   &  0   &  0  &   1&   0\\
0  &   0   &  0   &  0  &   0&   1\\
\end{array}
\right].
\end{eqnarray*}

By Definition 2.5, we obtain
\begin{eqnarray*}
\mathcal {X}_{XH(X)}&=&\prod(\mathscr{C}^{+})\cdot \mathcal {X}_{X}
=\left[
\begin{array}{ccccccc}
1  &   0   &  0   &  0  &   0&   0\\
1  &   1   &  0   &  1  &   1&   0\\
0  &   0   &  1   &  0  &   0&   0\\
0  &   0   &  0   &  1  &   0&   0\\
0  &   0   &  0   &  0  &   1&   0\\
0  &   0   &  0   &  0  &   0&   1\\
\end{array}
\right] \cdot \left[
\begin{array}{c}
0 \\
0 \\
1 \\
1 \\
1 \\
0\\
\end{array}
\right] =\left[
\begin{array}{cccccc}
0 & 1 & 1 & 1 & 1 & 0 \\
\end{array}
\right]^{T};\\
 \mathcal {X}_{XL(X)}&=&\prod(\mathscr{C}^{+})\odot
\mathcal {X}_{X}=\left[
\begin{array}{ccccccc}
1  &   0   &  0   &  0  &   0&   0\\
1  &   1   &  0   &  1  &   1&   0\\
0  &   0   &  1   &  0  &   0&   0\\
0  &   0   &  0   &  1  &   0&   0\\
0  &   0   &  0   &  0  &   1&   0\\
0  &   0   &  0   &  0  &   0&   1\\
\end{array}
\right] \odot \left[
\begin{array}{c}
0 \\
0 \\
1 \\
1 \\
1 \\
0\\
\end{array}
\right]=\left[
\begin{array}{ccccccc}
0 & 0 & 1 & 1 & 1 & 0 \\
\end{array}
\right]^{T}.
\end{eqnarray*}

Therefore, $XH(X)=\{x_{2},x_{3},x_{4},x_{5}\}$ and
$XL(X)=\{x_{3},x_{4},x_{5}\}$.

In Example 3.5, we need to compute all elements in
$\prod(\mathscr{C}^{+})$ for constructing approximations of sets by
Definition 3.1. By Theorem 3.4, we only need to calculate elements in
$\triangle_{1}(\prod(\mathscr{C}))$,
$\triangle_{2}(\prod(\mathscr{C}))$, $\triangle_{3}(\prod(\mathscr{C}))$
and $\triangle_{4}(\prod(\mathscr{C}))$. Thereby, the incremental algorithm is more effective
to compute approximations of sets.

\section{Non-incremental and incremental algorithms for computing the second and sixth lower and upper approximations of sets}

In this section, we show non-incremental and incremental algorithms of computing the second lower and
upper approximations of sets.

\begin{algorithm}(Non-incremental algorithm of computing $SH_{\mathscr{C}^{+}}(X^{+})$ and $SL_{\mathscr{C}^{+}}(X^{+})\bf{(NCS)}$)

Step 1: Input $(U^{+},\mathscr{C}^{+})$ and $X^{+}\subseteq
U^{+}$;

Step 2: Construct $M_{\mathscr{C}^{+}}$ and
$\Gamma(\mathscr{C}^{+})=M_{\mathscr{C}^{+}}\cdot M^{T}_{\mathscr{C}^{+}};$

Step 3: Compute $\mathcal {X}_{SH_{\mathscr{C}^{+}}(X^{+})}=\Gamma(\mathscr{C}^{+})\cdot \mathcal
{X}_{X^{+}}$ and $\mathcal {X}_{SL_{\mathscr{C}^{+}}(X^{+})}=\Gamma(\mathscr{C}^{+})\odot \mathcal
{X}_{X^{+}}$;

Step 4: Output $SH_{\mathscr{C}^{+}}(X^{+})$ and $SL_{\mathscr{C}^{+}}(X^{+})$.
\end{algorithm}

\begin{algorithm}(Incremental algorithm of computing $SH_{\mathscr{C}^{+}}(X^{+})$ and $SL_{\mathscr{C}^{+}}(X^{+})\bf{(ICS)}$)

Step 1: Input $(U,\mathscr{C})$, $(U^{+},\mathscr{C}^{+})$ and $X\subseteq
U^{+}$;

Step 2: Calculate $\Gamma(\mathscr{C})=M_{\mathscr{C}}\cdot
M^{T}_{\mathscr{C}}$, where $ M_{\mathscr{C}}=(a_{ij})_{n\times m};
$

Step 3: Compute $\triangle_{1}(\Gamma(\mathscr{C}))$ and
$\triangle_{2}(\Gamma(\mathscr{C}))$  and  $\triangle_{3}(\Gamma(\mathscr{C}))$;

Step 4: Construct $\Gamma(\mathscr{C}^{+})$, where
\begin{eqnarray*}
 \Gamma(\mathscr{C}^{+})&=&(c_{ij})_{(n+1)(n+1)}=\left[
  \begin{array}{cc}
    \Gamma(\mathscr{C}) & 0\\
    0 & 0\\
  \end{array}
\right]\bigvee \left[
  \begin{array}{cc}
    \triangle_{1}(\Gamma(\mathscr{C})) & (\triangle_{2}(\Gamma(\mathscr{C})))^{T}\\
    \triangle_{2}(\Gamma(\mathscr{C})) & \triangle_{3}(\Gamma(\mathscr{C}))\\
  \end{array}
\right];
\end{eqnarray*}

Step 5:  Obtain $\mathcal {X}_{SH(X)}$ and $\mathcal
{X}_{SL(X)}$, where
\begin{eqnarray*}
\mathcal {X}_{SH(X)}&=&\Gamma(\mathscr{C}^{+})\cdot \mathcal
{X}_{X}; \mathcal {X}_{SL(X)}=\Gamma(\mathscr{C}^{+})\odot \mathcal
{X}_{X}.
\end{eqnarray*}
\end{algorithm}

Subsequently, we present non-incremental and incremental algorithms of computing the sixth lower and
upper approximations of sets.

\begin{algorithm}(Non-incremental algorithm of computing $XH_{\mathscr{C}^{+}}(X^{+})$ and $XL_{\mathscr{C}^{+}}(X^{+})\bf{(NCX)}$)

Step 1: Input $(U^{+},\mathscr{C}^{+})$ and $X^{+}\subseteq
U^{+}$;

Step 2: Construct $M_{\mathscr{C}^{+}}$ and
$\prod(\mathscr{C}^{+})=M_{\mathscr{C}^{+}}\cdot M^{T}_{\mathscr{C}^{+}};$

Step 3: Compute $\mathcal {X}_{XH_{\mathscr{C}^{+}}(X^{+})}=\prod(\mathscr{C}^{+})\cdot \mathcal
{X}_{X^{+}}$ and $\mathcal {X}_{XL_{\mathscr{C}^{+}}(X^{+})}=\prod(\mathscr{C}^{+})\odot \mathcal
{X}_{X^{+}}$;

Step 4: Output $XH_{\mathscr{C}^{+}}(X^{+})$ and $XL_{\mathscr{C}^{+}}(X^{+})$.
\end{algorithm}

\begin{algorithm}(Incremental algorithm of computing $XH_{\mathscr{C}^{+}}(X^{+})$ and $XL_{\mathscr{C}^{+}}(X^{+})\bf{(ICX)}$)

Step 1: Input $(U,\mathscr{C})$, $(U^{+},\mathscr{C}^{+})$ and $X\subseteq
U^{+}$;

Step 2: Construct $\prod(\mathscr{C})$, where
$\prod(\mathscr{C})=M_{\mathscr{C}}\odot M^{T}_{\mathscr{C}};$

Step 3: Compute $\triangle_{1}(\prod(\mathscr{C}))$ and
$\triangle_{2}(\prod(\mathscr{C}))$, $\triangle_{3}(\prod(\mathscr{C}))$ and $\triangle_{4}(\prod(\mathscr{C}))$;

Step 4: Calculate $\prod(\mathscr{C}^{+})$, where
 $\prod(\mathscr{C}^{+})=\left[
  \begin{array}{cc}
    \prod(\mathscr{C}) & 1\\
    1 & 1\\
  \end{array}
\right]\bigwedge\left[
  \begin{array}{cc}
    \triangle_{1}(\prod(\mathscr{C})) & \triangle_{3}(\prod(\mathscr{C}))\\
    \triangle_{2}(\prod(\mathscr{C})) & \triangle_{4}(\prod(\mathscr{C}))\\
  \end{array}
\right];$

Step 5:  Get $\mathcal {X}_{XH(X)}$ and $\mathcal {X}_{XL(X)}$,
where
\begin{eqnarray*}
\mathcal {X}_{XH(X)}&=&\prod(\mathscr{C}^{+})\cdot \mathcal {X}_{X};
\mathcal {X}_{XL(X)}=\prod(\mathscr{C}^{+})\odot \mathcal {X}_{X}.
\end{eqnarray*}
\end{algorithm}

\section{Conclusions}

In this paper, we have provided effective approaches to constructing
approximations of concepts in dynamic covering approximation spaces.
Concretely, we have constructed type-1 and type-2 characteristic
matrices of coverings with the incremental approaches. Incremental
algorithms have been presented for computing the second and sixth
lower and upper approximations of sets. Several examples have been
employed to illustrate that computing approximations of sets could
be reduced greatly by using the incremental approaches.

In the future, we will propose more effective approaches to
constructing the type-1 and type-2 characteristic matrices of
coverings. Additionally, we will focus on the development of
effective approaches for knowledge discovery in dynamic covering
approximation spaces.

\section*{ Acknowledgments}

We would like to thank the anonymous reviewers very much for their
professional comments and valuable suggestions. This work is
supported by the National Natural Science Foundation of China (NO.
11201490,11371130,11401052,11401195), the Scientific Research Fund of Hunan
Provincial Education Department(No.14C0049).

\end{document}